# A Feature Selection Method for Driver Stress Detection Using Heart Rate Variability and Breathing Rate

Ashkan Parsi, David O'Callaghan, Joseph Lemley
*OCTO Sensing Team, Xperi Corporation,* Galway, Ireland
{ashkan.parsi, david.ocallaghan, joe.lemley}@xperi.com

## ABSTRACT

Driver stress is a major cause of car accidents and death worldwide. Furthermore, persistent stress is a health problem, contributing to hypertension and other diseases of the cardiovascular system. Stress has a measurable impact on heart and breathing rates and stress levels can be inferred from such measurements. Galvanic skin response is a common test to measure the perspiration caused by both physiological and psychological stress, as well as extreme emotions. In this paper, galvanic skin response is used to estimate the ground truth stress levels. A feature selection technique based on the minimal redundancy-maximal relevance method is then applied to multiple heart rate variability and breathing rate metrics to identify a novel and optimal combination for use in detecting stress. The support vector machine algorithm with a radial basis function kernel was used along with these features to reliably predict stress. The proposed method has achieved a high level of accuracy on the target dataset.

## 1. INTRODUCTION

Any type of action considered as a threat that could lead to the sympathetic nervous system (SNS) response, will cause stress. This act generates a sudden release of different hormones such as adrenaline and cortisol in the body [1]. Automobile drivers would be affected by this condition which could lead to undesirable events such as lane departure, running red lights, traffic noise, traffic congestion, and symptoms such as lack of sleep, driving phobia, impatience, and fatigue [2], [3]. All of these can limit a driver's concentration and ability to make rational decisions, which would increase the risk of car accidents. Considering stress is one of the major causes of automobile crashes each year [4], having a reliable stress detection system in an advanced driver-assistance system (ADAS) would be crucial and advantageous to saving lives.

Given that stress is regulated by the autonomous nervous system, it can be measured via other physiological measurements such as blood pressure, galvanic skin response (GSR), electromyogram (EMG), breathing frequency, respiration rate (RESP), and temperature [5]. Heart rate variability (HRV) as a measure of the balance between sympathetic mediators of heart rate (HR) (i.e. the effect of epinephrine and norepinephrine, released from sympathetic nerve fibers [6]) is one of these physiological measurements [7]. HRV operates on different time scales to help us adapt to environmental and psychological challenges. It also reflects the regulation of autonomic balance, blood pressure, heart, and vascular tone, which all could be affected by stress [7]–[9].

In this study, instantaneous heart rate (IHR) and respiration rate have been used as input to the stress detection system to detect stress labels which are calculated based on GSR values. This work proposes a new feature selection method to optimize the number of metrics used in detecting stress based on the input signals. The layout of the paper is as follows. The database used, extracted features, and feature selection method are outlined in Section II. Section III describes the implemented signal pre-processing and machine learning methods along with results and finally, conclusions are drawn in Section IV.

## 2. MATERIAL AND METHODS

### 2.1. Database

To develop an applicable stress predictive model from physiological signals, the first requirement is an accurately acquired database based on real scenarios that happen for the driver. The dataset used in this study was MIT "drivedb" [10], a dataset collected at MIT by Healey and Picard [11], and a part of the database is publicly available on PhysioNet (more precisely, datasets related to 17 experiments are released publicly) [12]. It contains data from 20 miles of driving in the greater Boston area, USA. The dataset consists of a collection of multi-parameter recordings obtained from 9 young and healthy individuals while they were driving on a designated route in the city and highways around Boston, Massachusetts. The driving protocol was as follows:

- The rest periods were used to gather baseline measurements and to create a low-stress situation.
- Drivers drove through side streets until they reached a busy main street in the city. The high-stress situation, where the drivers encountered stop-and-go traffic and had to contend with unexpected hazards such as cyclists and pedestrians.
- The route then led drivers away from the city, over a bridge, and onto a highway to make the medium-stress condition.
- The drivers then re-entered the highway heading in the opposite direction to repeat the above protocol.

The following physiological signals were recorded from the designed experiment: EDA measured in two placements (hand and foot), electrocardiogram (ECG), IHR, EMG, and respiration rate. The stress metric proposed in this work [11] validated the assumption of the low, medium, and high-stress levels during the rest, highway, and city driving, respectively. However, as stress can have several sources depending on the mental state of the participants [13], [14], the hand GSR signal has been used to label the stress in this study. The median GSR values have been used as the cut-off point in previous studies [4]. In this work to train the method on different levels of stress, $median \pm \alpha$ has been considered the cut-off point. Therefore, values above the $median + \alpha$ were labelled as stress while the values below the $median - \alpha$ were labelled as no stress. After running an optimization step $\alpha = \sigma/2$ where $\sigma$ is the standard deviation of the GSR signal.

## 2.2. HRV Features Extraction

HRV is defined as the physiological variation in the duration of intervals between sinus beats. It serves as a measurable indicator of cardiovascular integrity. HRV reflects both the sympathetic and the parasympathetic components of the autonomic nervous system which could be affected by stress and it has been proven that HRV decreases as stress increases [6], [15]. The IHR signal is calculated from the R-R intervals recorded using the following equation:

$$IHR\ (\text{bpm}) = \frac{60{,}000}{RRI\ (\text{ms})} \quad (1)$$

Where *RRI* (R-R intervals) is the time in milliseconds between instantaneous heartbeats as measured by any ECG wearable sensor, or the period between adjacent QRS complexes resulting from sinus node depolarizations, which is termed the normal to normal (NN) interval [6]. The unit of IHR is beats/minute (bpm).

After ectopic beats have been removed using the median filter based on the method proposed in [16], NN intervals are passed to the feature extraction step. Generally, the major features in HRV analysis can be divided into four main groups based on the study in 1996 [17] which was updated later in 2015 [18] by a joint task force between the European Society of Cardiology and the North American Society of Pacing and Electrophysiology. These groups are described as follows:

### 2.2.1. Time Domain Analysis

Standard time domain parameters of HRV analysis are the simplest to extract and are valuable measures of variation of the R-R intervals [17]. The most used time domain features are presented in Table I. The mean and standard deviation of the NN intervals are the most common time domain features. Other time domain features include the mean square of successive NN intervals differences and the number of adjacent NN intervals differing by more than 20 and 50 milliseconds (NN20, NN50). pNN20 and pNN50, the numbers obtained by dividing NN20 and NN50 respectively by the total number of NN intervals, have also been calculated. HRV triangular index (HRVTri), energy, zero crossing, and features from NN interval autocorrelation on 30-second windows also have been considered to study the effect of stress on NN intervals.

### 2.2.2. Frequency Domain Analysis

Isolation of the sympathetic activity which is associated with the low-frequency modulation of the heart rate (0.04-0.15 Hz) and parasympathetic activity which is associated with the higher frequency modulation (0.15-0.4 Hz) is the focus of the frequency domain analysis. Many studies have used the power spectral density (PSD) estimate in the frequency range of interest, such as very low frequency (VLF) band (0.0-0.04 Hz), low frequency (LF) band (0.04-0.15 Hz), high frequency (HF) band (0.15-0.4 Hz) as presented in Table I.

TABLE I. EXTRACTED FEATURES IN TIME DOMAIN, FREQUENCY DOMAIN, BISPECTRUM (31 FEATURES), NONLINEAR AND RESPIRATION RATE ANALYSIS

| **Features Category** | **Features Name** |
|---|---|
| **Time domain analysis (12 features)** | MeanNN, SDNN, and MSE |
| | NN20, NN50, pNN20 and pNN50 |
| | HRVTri, which is a total number of NN intervals used in histogram analysis divided by the maximum of the signal histogram, is employed in HRV analysis to extract, and evaluate geometrical feature. |
| | NN interval Energy, mean of zero-crossing, mean and minimum of NN interval autocorrelation |
| **Frequency domain analysis (14 features)** | PSD in VLF, LF, HF, ROI, LF/HF ratio, and Peak HF (peak of the highest frequency band) |
| | Relative, Logarithmic, and Normalized Power in VLF, LF, and HF |
| **Bispectrum analysis (31 features)** | $M_{avg}$ in LL, LH, HH, and ROI (magnitude average of bispectrum) |
| | $P_{avg}$ in LL, LH, HH, and ROI (power average of bispectrum) |
| | $E_{nb}$ in LL, LH, HH, and ROI (normalized bispectrum entropy) |
| | $E_{snb}$ in LL, LH, HH, and ROI (squared normalized bispectrum entropy) |
| | $L_m$ in LL, LH, HH, and ROI (sum of the logarithmic magnitude of bispectrum) |
| | $L_{dm}$ in LL, HH, and ROI (sum of the logarithmic magnitude of the diagonal elements of bispectrum) |
| | $WCOB_i$ in LL, LH, HH, and ROI (represent the weight center of the contour map of the bispectrum – first frequency index) |
| | $WCOB_j$ in LL, LH, HH, and ROI (represent the weight center of the contour map of the bispectrum – second frequency index) |
| **Nonlinear analysis (5 features)** | $SD_1$, $SD_2$, $SD_1/SD_2$ ratio |
| | Rényi entropy and Tsallis entropy |
| **Respiration analysis (14 features)** | Mean$_{RESP}$, Median$_{RESP}$, SD$_{RESP}$, Max$_{RESP}$, and Min$_{RESP}$ |
| **$P_1$ = 0.06 Hz, $P_2$ = 0.12 Hz, $P_3$ = 0.24 Hz,** | PSD in $P_{12}$, $P_{23}$, $P_{34}$, $P_{45}$, $P_{56}$, $P_{67}$ |
| **$P_4$ = 0.49 Hz, $P_5$ = 0.98 Hz, $P_6$ = 1.95 Hz,** | Relative Power (PSD$_{P45}$ + PSD$_{P56}$ + PSD$_{P67}$) / (PSD$_{P12}$ + PSD$_{P23}$ + PSD$_{P34}$) |
| **$P_7$ = 3.91 Hz** | Sum of Powers and Max Power / Sum of Powers ratio |

LL: LF–LF where LF is low frequency band (0.04-0.15 Hz)
LH: LF–HF where LF is low frequency band (0.04-0.15 Hz), and HF is high frequency (HF) band (0.15-0.4 Hz)
HH: HF–HF where HF is high frequency (HF) band (0.15-0.4 Hz)
$SD_1$, $SD_2$: the Poincaré plot derived from HRV data represents each R-R interval as a function of the previous R-R interval. The width ($SD_1$) of generated ellipse, and the length ($SD_2$) of it can be calculated from this plot presenting valuable information for cardiac signals.
RESP: respiration rate

#### 2.2.3. Bispectrum Analysis

Higher-order spectral features up to third-order cumulant were employed to estimate the bispectrum from HRV data. Since the HRV signal is nonlinear and non-Gaussian by nature [19], [20], the bispectrum can be used to reveal information not present in the spectral domain. There are three sub-bands inside the region of interest (ROI) of the bispectrum which can discriminate sympathetic and parasympathetic contents of HRV signals [21], [22] which have been used to extract the features presented in Table I.

#### 2.2.4. Nonlinear Analysis

Viewing HRV as an indirect electrical measure of autonomic heart rate modulations, as the output of a nonlinear system, can lead to a better understanding of cardiac system dynamics [23]–[25]. Studies have also stressed the importance of nonlinear techniques to study HRV in issues related to both cardiovascular health and disease [14], [26]. Commonly used features in the nonlinear category have been used as described in Table I.

### 2.3. Respiration Analysis

Many studies show the response of the respiratory system and the increase in the respiratory rate during mental stress [27]. In this study, by looking at the most common breathing rate frequencies, several features have been extracted as presented in Table I.

TABLE II. SELECTED FEATURES ON EACH STEP IN THE PROPOSED FEATURE SELECTION PROCESS

| **First Step** | |
|---|---|
| **Features Category** | **Features Name in mRMR order** |
| Time domain analysis<br>(5 selected features) | HRVTri, pNN20, Min of autocorrelation, mean of zero crossing, and NN interval energy |
| Frequency domain analysis<br>(7 selected features) | PSD in HF,<br>Logarithmic Power in HF, LF, and ROI<br>PSD in LF, and ROI, Peak HF |
| Bispectrum analysis<br>(28 features) | $L_m$ in LH, ROI, LL, $M_{avg}$ in LH, $L_m$ in HH, $L_{dm}$ in HH, LL, $M_{avg}$ in HH, $L_{dm}$ in ROI, $P_{avg}$ in ROI, $M_{avg}$ in ROI, $P_{avg}$ in LL, LH, and HH, $WCOB_i$ in ROI, HH, and LL, $WCOB_j$ in LL, $WCOB_i$ in LH, $WCOB_j$ in LH, $E_{snb}$ in ROI, $E_{nb}$ in LH, LL, $E_{snb}$ in HH, $WCOB_j$ in ROI and HH, $E_{nb}$ in ROI, HH |
| Nonlinear analysis<br>(4 features) | $SD_2$, $SD_1/SD_2$, Rényi entropy and Tsallis entropy |
| Respiration analysis<br>(13 features) | $Med_{RESP}$, $Max_{RESP}$, PSD in $P_{56}$, Sum of Powers, Relative Power, $SD_{RESP}$, PSD in $P_{23}$, $Min_{RESP}$, Max Power / Sum of Powers, PSD in $P_{67}$, PSD in $P_{12}$, PSD in $P_{34}$, PSD in $P_{45}$ |
| **Second Step** | |
| All Categories<br>(20 feature) | Peak HF, $Max_{RESP}$, $WCOB_i$ in LL and HH<br>$WCOB_j$ in LH, $WCOB_i$ in LH, $WCOB_j$ in LL, $SD_1$, $L_m$ in HH, $WCOB_j$ in ROI, $E_{snb}$ in HH, PSD in $P_{56}$, Relative Power, PSD in $P_{45}$, Tsallis entropy, $Med_{RESP}$, $E_{snb}$ in LL, $Min_{RESP}$, $SD_{RESP}$, $E_{snb}$ in ROI |
| **Final Set**<br>**(18 selected features)**<br>The final feature set is the intersection of all categories which each individual one | $Max_{RESP}$, $WCOB_i$ in LL and HH, $WCOB_j$ in LH, $WCOB_i$ in LH, $WCOB_j$ in LL, $L_m$ in HH, $WCOB_j$ in ROI, $E_{snb}$ in HH, PSD in $P_{56}$, Relative Power, PSD in $P_{45}$, Tsallis entropy, $Med_{RESP}$, $E_{snb}$ in LL, $Min_{RESP}$, $SD_{RESP}$, $E_{snb}$ in ROI |

## 2.4. SVM Classifier

A support vector machine (SVM) classifier was applied to detect stressful and non-stressful episodes. Historically, SVMs have shown to be an effective multi-class discriminator [28] and previous studies have highlighted its efficacy with HRV signal classification [29]. The main idea of SVMs is to map the input data from the N-dimensional space, through some nonlinear mapping, to the M-dimensional feature space M > N [30] and this step happens using the help of a kernel function. In this study, the radial basis function has been used as a kernel with the L2 regularization parameter set to 4 and the scale set to 0.5. To get these values, the Bayesian optimization algorithm has been used.

## 2.5. Leave-one-out Cross-validation Method

To build a good learning model for feature selection and dealing with noise in the data, a cross-validation solution has been used [30]. This allows a proportion $(K-1)/K$ of the available experiment to be used for training while making use of all the data to assess performance. In this study the leave-one-out technique has been implemented to select features from each feature's categories (time, frequency, bispectrum, nonlinear, and respiration rate analysis).

## 2.6. Feature selection method

The minimal redundancy-maximal relevance method (mRMR) was implemented to rank the features according to their maximal statistical dependency on the target class [31], [32] for the features listed in Table I. However, instead of using the method on the whole feature set (as proposed in [32]), the top features from each category in time, frequency, bispectrum, nonlinear, and respiration rate analysis were calculated in the first step utilizing the mRMR feature ranking method. In the second step, the top-ranked features have been selected while the feature selection method has been applied to all feature categories together as one set. To make the final set, the common features from the first and the second step are selected. Features that are presented as top-rank metrics are not only the best of their category but also have been observed while the whole set has been tested to reduce redundancy even more. The results of each step are presented in Table III and Figure 1 is illustrating the feature selection method utilizing leave-one-out cross-validation performances on different features categories.

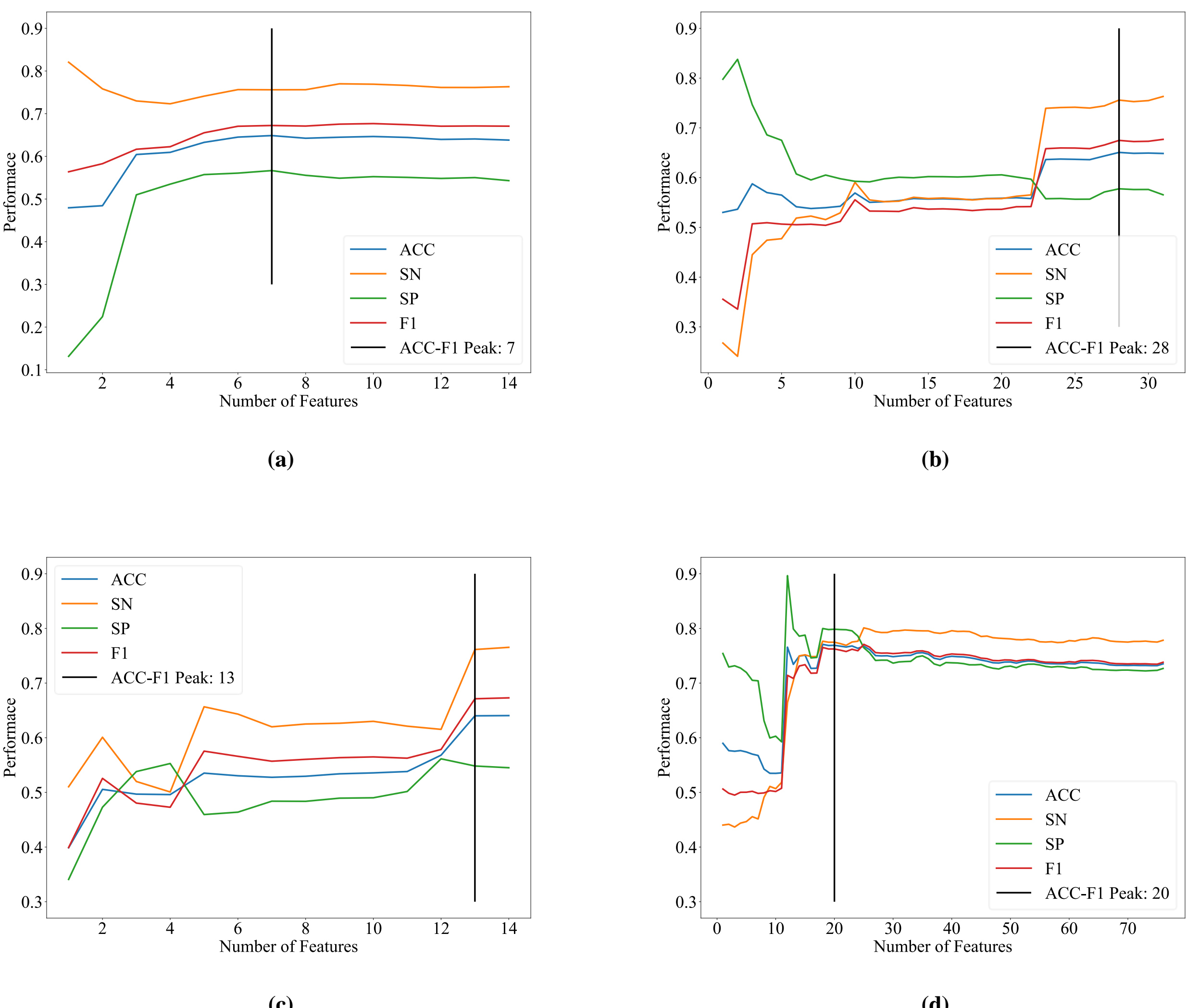

Figure. 1. Leave-one-out cross-validation performance of the different feature-domain categories including (a) frequency, (b) bispectrum, (c) respiration, and (d) all domains by adding them based on mRMR ranking.

ACC: accuracy
SN: sensitivity
SP: specificity
F1: F1 score

Figure. 2.

## 3. RESULTS

The dataset is split into two, non-overlapping subsections. This is done to ensure there is a robust evaluation of the feature selection technique. The first subset of data is used to rank and select the appropriate features in the time domain, frequency domain, bispectrum, nonlinear, and respiration analysis. The second subset is then used to evaluate models using features chosen for each category and the final subset from the first section is for the purpose of evaluation. Results from these experiments are presented in Table III.

When only using individual categories for feature selection, the best results were obtained by respiration analysis followed by time domain analysis. For 13 respiration features the best result was 93.8% in sensitivity and 81.8% in specificity. With a total number of 76 features, the best result was 95.5% in sensitivity and 75.2% in specificity. However as presented in Table I, the best result, with 93.3% in sensitivity and 83.9% in specificity, was achieved using 18 selected features. Out of these 18 features, seven were selected from respiration analysis, ten from bispectrum analysis, and one was selected from the nonlinear analysis.

TABLE III. SUMMARIES OF EXPERMENTAL RESULTS CONDUCTED USING DIFFERET TYPE OF FEATURES

| **Features Category** | **ACC (%)** | **SN (%)** | **SP (%)** | **F1 (%)** |
|---|---|---|---|---|
| Time domain analysis (5 selected features) | 69.2 | 72.2 | 67 | 66.9 |
| Frequency domain analysis (7 selected features) | 50.7 | 42.2 | 57.2 | 42.4 |
| Bispectrum analysis (28 features) | 46 | 66.3 | 30.8 | 51.4 |
| Nonlinear analysis (4 features) | 59.8 | 24.3 | 86.7 | 34.3 |
| Respiration analysis (13 features) | 87 | 93.8 | 81.8 | 86.1 |
| All Categories (20 feature) | 86.5 | 94 | 80.9 | 85.7 |
| All Feature without feature selection (76 features) | 84 | **95.5** | 75.2 | 83.7 |
| **Final Set (18 selected features)** | **87.9** | 93.3 | **83.9** | **86.9** |

ACC: accuracy
SN: sensitivity
SP: specificity
F1: F1 score

# 4. CONCLUSION

In this work, we present a method to select the most effective HRV and respiration features for detecting stress. Using the mRMR ranking method combined with the SVM classifier 18 features have been selected. We found that stress not only increases respiratory rate but also shows more changes in higher-order spectral features presented by bispectrum LL, LH, and HH frequency powers. Using this combination of features, the proposed method provides an improvement of over 6% in accuracy over the leading published results in the literature using the same classifier (87.9% vs 82%) and over 8% in the F1 score (86.9% vs 78%) [4]. These results demonstrate that having a stress detection system within an ADAS remains a challenging task as the input of the current system, however high in performance, uses wearable (invasive) sensors input such as ECG and respiration signals.